\documentclass{article}

\usepackage{PRIMEarxiv}

\usepackage{amsmath}
\usepackage{microtype}
\usepackage{xcolor}
\usepackage{hyperref}
\usepackage{url}
\usepackage{booktabs}
\usepackage[margin=1in]{geometry}
\usepackage{courier} 
 
\usepackage{lineno}
\usepackage{graphicx}
\usepackage{subcaption}
\usepackage{enumitem}
\usepackage{natbib}

\definecolor{darkblue}{rgb}{0, 0, 0.5}
\hypersetup{colorlinks=true, citecolor=darkblue, linkcolor=darkblue, urlcolor=darkblue}

\newcommand{\fakepar}[1]{\vspace{1mm}\noindent\textbf{#1:}}

\title{TweakLLM: A Routing Architecture for Dynamic Tailoring of Cached Responses}

\date{\today}

\author{
  Muhammad Taha Cheema, Abeer Aamir \& Khawaja Gul Muhammad  \thanks{Equal contribution of all primary authors.}\\
  Department of Computer Science\\
    Lahore University of Management Sciences\\
    Lahore, Pakistan\\
  \texttt{\{25100203,25100161,25100307\}@lums.edu.pk} \\ \AND
  Naveed Anwar Bhatti, Ihsan Ayyub Qazi \& Zafar Ayyub Qazi \\
    Department of Computer Science \\
    Lahore University of Management Sciences \\
    Lahore, Pakistan \\
    \texttt{\{naveed.bhatti,ihsan.qazi,zafar.qazi\}@lums.edu.pk} 
    }

\begin{document}
\maketitle

%%%%%%%%%%%%%%%%%%%%%%%%%%%%%%%%%%%%%%%%%%%
\begin{abstract}
Large Language Models (LLMs) process millions of queries daily, making efficient response caching a compelling optimization for reducing cost and latency. However, preserving relevance to user queries using this approach proves difficult due to the personalized nature of chatbot interactions and the limited accuracy of semantic similarity search. To address this, we present TweakLLM, a novel routing architecture that employs a lightweight LLM to dynamically adapt cached responses to incoming prompts. Through comprehensive evaluation—including user studies with side-by-side comparisons, satisfaction voting, as well as multi-agent LLM debates—we demonstrate that TweakLLM maintains response quality comparable to frontier models while significantly improving cache effectiveness. Our results across real-world datasets highlight TweakLLM as a scalable, resource-efficient caching solution for high-volume LLM deployments without compromising user experience.
\end{abstract}

%%%%%%%%%%%%%%%%%%%%%%%%%%%%%%%%%%%%%%%%%%%
\section{Introduction}
Modern LLM systems are responsible for handling millions of queries every day. As the cost of running models with a large number of parameters increases, providers are motivated to optimize inference pipelines~\citep{brown2023gpt4,microsoft2023fy23,pichai2023google,touvron2023llama2}. One particularly promising direction is caching, which serves stored responses to similar queries, reducing redundant computations and thereby cutting both operational costs and response latency~\citep{bang-2023-gptcache,regmi-2024-gpt-semantic-cache,aws_semantic_cache}.

However, caching in the context of personalized, free-form dialogue is challenging. Semantic caching uses embedding models to assess the similarity of prompts. Yet, even small changes in wording can reflect significant shifts in user intent, making it risky to reuse responses based on embedding similarity alone. This is key given that one of the major strengths of LLMs is their ability to generate highly contextualized and personalized responses, a capability that renders precision critical for real-world use of any caching mechanism.

Prior work using such a caching framework identified the issue of precision. \citep{bang-2023-gptcache}. At low similarity thresholds, false positive cache hits occur at rates unacceptable for production environments, while at higher thresholds, cache utilization became negligible. This observation has been corroborated by \cite{regmi-2024-gpt-semantic-cache}, who found similar precision-recall tradeoffs in large-scale deployment settings. This makes semantic similarity alone insufficient for effective caching in practical LLM systems, necessitating a more sophisticated approach.

In light of this, we propose \textit{TweakLLM}.\footnote{Our work is open-source. The complete TweakLLM framework and evaluation pipeline can be found at our  \href{https://github.com/kvcache842/tweak-llm}{GitHub repository}} Our approach introduces a two-tier system: First, a \textbf{\textit{semantic cache lookup}} where, when a user prompt arrives, we perform a standard semantic similarity search using a robust embedding model and a tunable cosine similarity threshold to retrieve candidate cached response(s). Second, \textbf{\textit{dynamic adjustment via a lightweight LLM}} where, rather than returning the top cached response verbatim, a small, cost-effective LLM is employed to tailor the cached response for relevance and accuracy to the nuances of the incoming prompt.

Our research makes the following key contributions:

\begin{enumerate}[leftmargin=1.4cm, label=\textbf{[C\arabic{enumi}]}]
    \item We empirically demonstrate that semantic similarity alone is fundamentally inadequate for LLM response caching, establishing quantitative boundaries of the precision-recall trade-off across multiple datasets and embedding models.
    
    \item We introduce \textit{TweakLLM}, a novel two-tier caching architecture that combines semantic lookup with lightweight response refinement, addressing the limitations of traditional caching approaches.
    
    \item We provide comprehensive evidence of TweakLLM's effectiveness through rigorous evaluation metrics, demonstrating significant cost reduction while maintaining response quality comparable to non-cached baseline.
\end{enumerate}

\fakepar{Paper Organization} The remainder of this paper is structured as follows: Section~\ref{sec:rw} situates our work within the broader literature; Section~\ref{sec:arch} details the TweakLLM architecture; Section~\ref{sec:exp} describes our experimental methodology, including datasets and evaluation protocols; Section~\ref{sec:eval} presents our empirical results; Section~\ref{sec:diss} discusses implications, limitations, and future directions; Section~\ref{sec:concl} concludes the paper.

%%%%%%%%%%%%%%%%%%%%%%%%%%%%%%%%%%%%%%%%%%%
\section{Related Work}
\label{sec:rw}

GPTCache \citep{bang-2023-gptcache} is a popular framework that utilizes a single-layer architecture whereby it returns cached responses for queries that have an embedding distance greater than a set threshold. Their architecture comprises embedding generation for the incoming prompt, cache lookup, and re-rank of the top cache hits through a cross-encoder to find the best candidate cached prompt. The corresponding cached response is then returned. They conclude that even with a suitable embedding model, positive
hit rates are unlikely to reach production requirements without decreasing cache hits. They further highlight another issue where existing embedding models may incorrectly match texts with opposite meanings but similar words, which is acceptable for search but fails for LLM caching scenarios. 

GPT Semantic Cache \citep{regmi-2024-gpt-semantic-cache} is based on a similar principle. On a custom-curated dataset, the authors report cache hits rates of up to 68.8\% and up to 97\% true positive hits. However, we believe that their evaluation methodology has significant limitations: they provide GPT4o-mini the test query, the corresponding cached query and the cached response, and ask it to respond in a binary verdict whether the the cached response is accurate for the test query.  For meaningful real-world application, the quality of a cached response should be measured \textit{with reference} to a frontier-model's response.  In TweakLLM's framework, we improve on their evaluations using multi-LLM debate inspired by ChatEval \citep{chan2023chatevalbetterllmbasedevaluators} as well as user surveys for more rigorous user experience testing. The authors of GPT Semantic Cache also highlight that domain-specific queries may require tailored embeddings for improved accuracy, which further motivates our tweaking architecture.   

In general, there is significant interest in the field of semantic search. AWS' read-through semantic cache architecture \citet{aws_semantic_cache} uses embeddings and adjustable thresholds to optimize LLM latency and cost. RouteLLM \citep{ong2025routellmlearningroutellms} introduces a learning framework to dynamically route queries between strong and weak LLMs using human preference data and data augmentation using LLMs, cutting costs by over 2x while preserving high-quality responses. We extend their routing approach to our caching framework. Accordingly, our work situates itself at the intersection of LLM routing and semantic caching, aiming to unify the adaptability of routing frameworks with the efficiency of cache-based retrieval.

Finally, there is a lot of interest in the potentials of using small LLMs (SLMs) for many tasks that do not require frontier models. \citet{microsoft_slm_function_calling} discusses how to fine-tune small LLMs for function calling. Studies indicate SLMs increasing viability for use in resource-efficient, domain-specific, and hybrid LLM systems \citep{wang2024comprehensivesurveysmalllanguage}. TweakLLM leverages this insight by employing a small LLM to refine cached responses. Our work shows that SLMs can bridge the gap between response quality and cache efficiency.

%%%%%%%%%%%%%%%%%%%%%%%%%%%%%%%%%%%%%%%%%%%
\section{System Architecture}
\label{sec:arch}

\begin{figure}[t]
  \centering
  \includegraphics[width=0.8\linewidth]{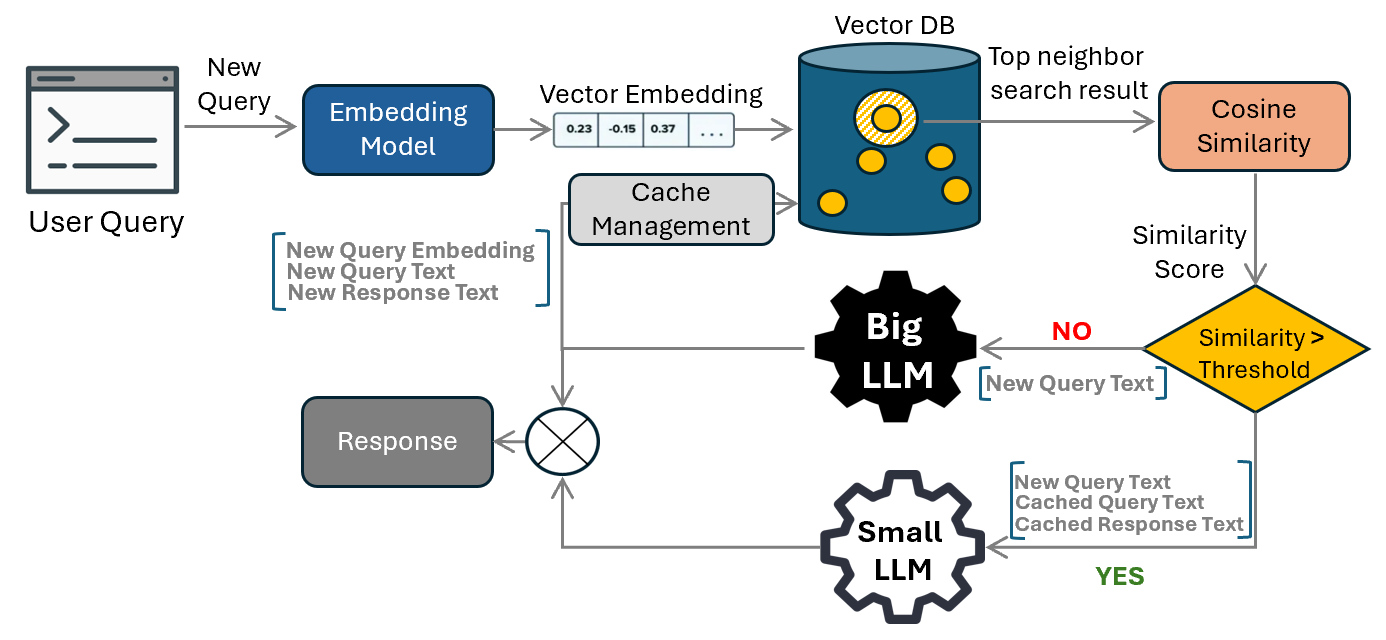}
  \caption{\textbf{\textit{TweakLLM} Architecture}: Incoming queries are embedded and compared to cached entries via cosine similarity. High similarity (above threshold) triggers response refinement by a Small LLM using the cached response. Low similarity results in generation by a Big LLM, with the new query-response pair being cached}
  \label{fig:simple-arch}
\end{figure}

\subsection{Overview}

TweakLLM employs a dual-model architecture, illustrated in Figure~\ref{fig:simple-arch}, designed to optimize the trade-off between response quality, latency, and computational cost. It combines a powerful, state-of-the-art large language model (\textbf{Big LLM}\footnote{We define Big LLM as a model with a large parameter count (e.g., GPT-4o, Llama 405B) and high API cost, used as a proxy for capability when parameter counts are undisclosed.}) with a smaller, faster counterpart (\textbf{Small LLM}\footnote{Models with significantly fewer parameters, lower computational needs, and reduced API/inference costs (e.g., Llama 3.1 8B, Qwen 2.5 7B).}). The core principle is leveraging semantic caching followed by targeted refinement: using the Small LLM to adapt previously generated high-quality responses for similar new queries, thereby reducing reliance on the more expensive Big LLM to generate responses afresh.

The system's operational workflow proceeds as follows:

    \fakepar{Query Embedding} Upon receiving a user query, an embedding model transforms the text into a dense vector representation capturing its semantic meaning.

    \fakepar{Cache Lookup and Similarity Evaluation} This query embedding is used to search a vector database, which stores embeddings, query texts, and corresponding response texts from previous Big LLM interactions. An Approximate Nearest Neighbor (ANN) search retrieves the cached entry with the highest cosine similarity to the new query.

    \fakepar{Threshold-Based Routing} The calculated similarity score is compared against a predefined threshold. This comparison dictates the subsequent processing path:

    \begin{itemize}
        \item \textbf{Cache Hit Pathway (Similarity $\ge$ Threshold):} The system retrieves the cached \texttt{query\_text} and \texttt{response\_text} associated with the top matching vector. These, along with the new \texttt{query\_text}, are passed to the \textbf{Small LLM}. The Small LLM is specifically prompted to refine the cached response, adapting it to address the nuances of the new query while preserving the core information and quality standard set by the Big LLM. This pathway significantly reduces latency and cost.

        \item \textbf{Cache Miss Pathway (Similarity $<$ Threshold):} The new query is routed directly to the \textbf{Big LLM} for comprehensive response generation from scratch.

    \end{itemize}

    \fakepar{Response Delivery and Cache Update} The generated or refined response is returned to the user. In the case of a cache miss, the new \texttt{query\_text}, its \texttt{query\_embedding}, and the freshly generated \texttt{response\_text} from the Big LLM are stored in the Vector Database via the cache management component. This continuously enriches the cache for potential future hits.

    The current implementation of cache management component utilizes a simple append-only caching strategy where every new query-response pair generated by the Big LLM is added to the vector database. The cache management component (Figure~\ref{fig:simple-arch}) is designed modularly, allowing for future extensions, as discussed in Section \ref{sec:diss}. 

%%%%%%%%%%%%%%%%%%%%%%%%%%%%%%%%%%%%%%%%%%%5
\section{Experimental Setup}
\label{sec:exp}

Appendix \ref{appendix:A:tweaker-prompt} shows the prompt used by Small LLM for tweaking cached responses. Table~\ref{tab:implementation-details} summarizes the specific components and configurations used in our TweakLLM implementation, on which we base our experimental analysis. 

\begin{table}[t] 
\centering
\begin{tabular}{@{}lp{0.65\linewidth}@{}} % Adjust p{} width as needed
\toprule
\textbf{Component / Parameter} & \textbf{Specification / Details} \\
\midrule
Big LLM & GPT-4o \citep{openai_gpt4o}. Chosen for state-of-the-art capabilities. \\
Small LLM & Llama 3.1 8B Instruct \citep{llama3.1_8b_instruct}. Offers significant cost reduction (approx. 25x less per output token than GPT-4o API). Generations for both done at default temperatures.\\
Embedding Model & \texttt{sentence-transformers/all-MiniLM-L6-v2} \citep{all_minilm_l6_v2}. Generates 384-dim embeddings optimized for semantic search. \\
Vector Database & Milvus v.2.5.x \citep{milvus}. Stores \texttt{query\_text}, \texttt{query\_embedding}, \texttt{response\_text}. Uses \texttt{IVF\_FLAT} index on embeddings for search acceleration. \\
Query Preprocessing & The phrase \textit{"answer briefly"} appended to all queries to encourage conciseness for evaluation purposes. \\
Initial Similarity Threshold & 0.7 (cosine similarity). Selected based on preliminary tests; further analysis performed across bands (0.7-0.8, 0.8-0.9, 0.9-1.0). \\
\bottomrule
\end{tabular}
\caption{TweakLLM implementation components and configuration}
\label{tab:implementation-details}
\end{table}

\subsection{Datasets} 
\label{sec:4.1-datasets}
We evaluate TweakLLM on a range of real-world datasets \citep{quora_dataset, lmsys-1m-dataset, wildchat-1m-dataset} selected to represent common use cases where caching can be advantageous:

\begin{itemize}
    \item \textbf{Question Pairs Dataset \citep{quora_dataset}}: This dataset comprises more than 400,000 pairs of related questions from Quora. Each pair is labeled as duplicate or non-duplicate by human annotators. This gives us a hand-picked subset of similar questions. 
    \item \textbf{allenai/WildChat-1M \citep{lmsys-1m-dataset} \& lmsys/lmsys-chat-1m \citep{wildchat-1m-dataset}:} We use these two open-source datasets which capture diverse LLM interactions, to assess performance on real-world conversational data. WildChat is a collection of 650K conversations with OpenAI's GPT-3.5 and GPT-4. Similarly, LMSYS contains 1 million user-chatbot conversations collected from conversations with 25 state-of-the-art LLM.
    
    We filter the LMSYS dataset to retain English conversations which are neither redacted nor flagged by OpenAI's moderation. Then, we pick only the first user query from each conversation. 
\end{itemize}

\subsection{Evaluation Methods}
Our evaluation setup consists of two main areas. First, we demonstrate the inefficacy of traditional semantic caching such as that used in GPTCache (Sec \ref{sec:4.2.1-eval-traditional}). Second, we demonstrate the effectiveness of our proposed TweakLLM architecture qualitatively (Sec \ref{sec:4.2.2-eval-tweakllm-qualitative}) and quantitatively (Sec \ref{sec:4.2.3-eval-tweakllm-quantitative}).

\label{sec:4.2.1-eval-traditional}
\subsubsection{Evaluation of Traditional Semantic Caching}
We use the Question Pairs Dataset and employ GPTCache's framework: We embed each question using sentence-transformers/all-MiniLM-L6-v2 \citep{all_minilm_l6_v2}. For each pair of similar questions in the dataset, we \texttt{put()} the first question and \texttt{get()} the second question of each pair. For the \texttt{get()} operation, the top-\textit{k} hits are retrieved using cosine distance. Then we use albert-duplicate-onnx~\citep{albert_duplicate_onnx} and quora-distilroberta-base~\citep{quora_distilroberta_base} to re-rank the candidates with respect to the original question in order to get the best match. After the \texttt{get()} operation, we also insert the second question into the cache via a \texttt{put()} call, enabling growth of the cache over time. We set different cosine thresholds for the vector database ANN / top-k retrieval and calculate the precision and recall of this system across each threshold. For the precision/recall metrics, we define the following:
\begin{itemize}
    \item True Positive (TP): A cache hit pair that is annotated as a duplicate by human judgment.
    \item False Positive (FP): A cache hit pair but is annotated as not a duplicate by human annotators. In such cases, returning a cached response would be inappropriate.
    \item False Negative (FN): A cache miss pair yet is annotated as a duplicate. This represents a missed opportunity for leveraging the cache.
\end{itemize}

\label{sec:4.2.2-eval-tweakllm-qualitative}
\subsubsection{TweakLLM Qualitative Evaluation}
We evaluate TweakLLM's performance using the Question Pairs Dataset and the LMSYS-Chat-1M dataset.

For the Question Pairs Dataset, we insert the first question from each of 2,000 labeled pairs into Milvus to simulate cache population. We then query Milvus using the second question from each pair, retrieving the top hit and its cosine similarity. This follows the TweakLLM pipeline, where an incoming query is routed to the Small LLM for tweaking if the similarity exceeds a predefined threshold (set to 0.7). For the filtered LMSYS dataset, we insert approximately half the queries (248,808 entries) and query 82,700 unseen prompts from the remaining portion, using the same threshold of 0.7.

In both datasets, since we aim to measure the \textit{quality} of tweaked responses, we evaluate only the subset of queries that result in cache hits—i.e., those routed to the Small LLM for response tweaking, as the rest would be handled directly by the Big LLM. To establish strong baselines, we also generate responses for these same queries directly from the Big LLM and directly from the Small LLM. Both LLMs use the default temperature settings during response generation.

For Question Pairs Dataset, our evaluation is:
\begin{enumerate}
    \item Big LLM (direct) vs. Small LLM Tweaked
    \item Big LLM (direct) vs. Small LLM Direct (no tweaking); serves as a control for our LLM-as-evaluators method. 
\end{enumerate}
For LMSYS, our evaluation is restricted to Big LLM Direct vs. Small LLM Tweaked.

Our evaluation goal is to understand the quality of responses for queries sent through the tweaking part of TweakLLM's pipeline. This is measure of user experience; stated simply, would someone using a chatbot powered by TweakLLM be satisfied with the quality of responses when cache hits occur and tweaked cached responses are returned from Small LLM instead of the regular generation from Big LLM. We do the qualitative analysis in the following ways:
\begin{enumerate}
    
    \item \textbf{User survey:} We select 120 queries from the Question Pairs Dataset (queried set), with 40 each from the cosine similarity bands 0.7-0.8, 0.8–0.9 and 0.9–1.0, and include them in our user survey. The evaluation compares responses from the Big LLM (direct generation) versus the Small LLM (tweaked cached responses). We also track the time-taken for users to fill the survey. Each participant is shown:
    \begin{itemize}
        \item 3 \textit{side-by-side comparison} questions: for a given prompt, participants are shown both the Big LLM and Small LLM (unlabelled and randomly shuffled order). They are asked: \textit{"For each question below, you will see two AI-generated responses. Pick the response you prefer"}. They are given the option to vote for one or the other or vote \textit{"I prefer both responses equally"}. This setup mirrors the side-by-side evaluation strategies adopted by Chatbot Arena \citep{zheng2023-chatbot-arena} and OpenAI's evaluation for new models on chatgpt.com. 

        \item 6 \textit{individual satisfaction rating} questions: participants are shown the original query along with just one response (either from Big LLM or Small LLM), and asked \textit{"For each question below, you will see an AI-generated response. If you received this response after asking the question from an LLM, would you rate it as satisfactory or not satisfactory?"}. They vote in binary. They are shown 3 queries with Small LLM responses and 3 with Big LLM responses. Their ordering if shuffled. 
    \end{itemize} 

    In our survey application, we distribute the queries shown to users as evenly as possible by dynamically picking those which have the least number of \textit{side-by-side} comparisons and \textit{satisfaction} votes so far. 
    
    \item \textbf{LLM-as-evaluator pipeline:} Since a user survey is unable to scale to more questions and datasets, we additionally use LLMs-as-evaluators. This involves a multi-agent debate of three GPT-4o personas (see Table \ref{tab:system-roles-in-LLM-debate}). We show each persona both the Big LLM response and Small LLM response blinded and ask it to reason about and rate the responses. It can either vote for response A or response B or vote AB indicating that both are of equal quality according to its criteria. Each agent's reasoning and rating is appended to the next agent, who is encouraged to engage with the previous personas. The prompt for each LLM evaluator can be found in Appendix \ref{appendix:B:prompts-for-llm-eval}. The debate takes place over two rounds and the majority verdict is considered. This method and the decision to use two rounds was inspired by ChatEval \citep{chan2023chatevalbetterllmbasedevaluators} whose authors suggest that  debate-based evaluation methods performs better than simple LLMs-as-a-jury evaluation methods. 
\end{enumerate}

\begin{table}[t]
\centering
\begin{tabular}{ll}
\toprule
\textbf{Agent} & \textbf{Primary Evaluation Focus} \\
\midrule
Factual Accuracy Evaluator & Truthfulness, logical consistency \\
User Experience Evaluator & Clarity, tone, expected user satisfaction \\
Relevance \& Completeness Evaluator & Answer coverage, alignment with question intent \\
\bottomrule
\end{tabular}
\caption{System roles in multi-agent LLM evaluation debate}
\label{tab:system-roles-in-LLM-debate}
\end{table}

\label{sec:4.2.3-eval-tweakllm-quantitative}
\subsubsection{TweakLLM Quantitative Evaluation}
Our goal is to estimate how many similar queries an LLM typically handles. For this, we utilize the LMSYS-Chat-1M and WildChat-1M datasets since these are large, real-world conversational corpora (see Section \ref{sec:4.1-datasets}). For each, we insert the first half of the dataset and then query the remaining half to see what proportion of the new queries are cache hits as we vary the cosine similarity threshold. We use this data to estimate the cost saving for LLM providers (based on API cost difference per token).

%%%%%%%%%%%%%%%%%%%%%%%%%%%%%%%%%%%%%%%%%%%
\section{Results}
\label{sec:eval}

\subsection{Standalone (Traditional) Semantic Caching}

\begin{figure}[t]
    \centering
    \includegraphics[width=0.6\linewidth]{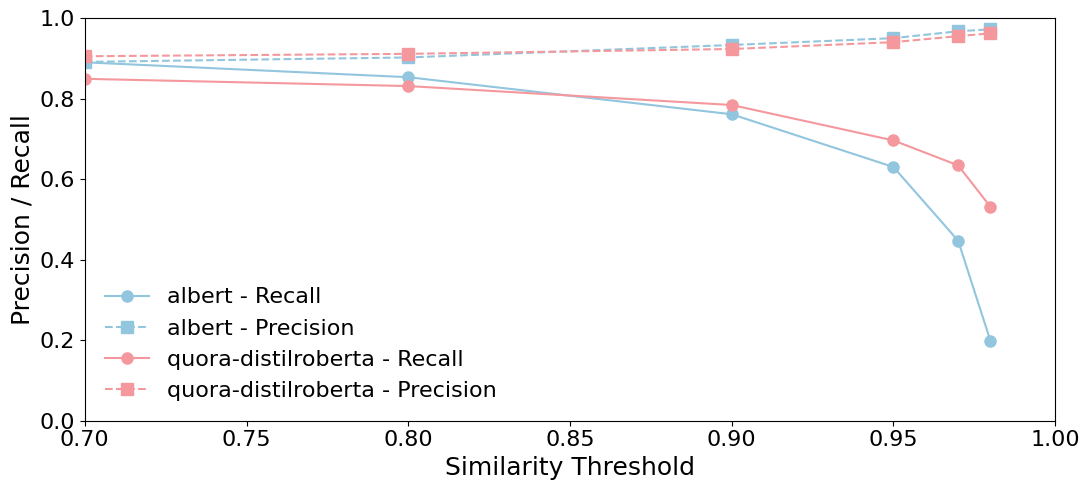}
    \caption{Precision-recall tradeoff while similarity thresholds on Question Pairs Dataset using GPTCache architecture. Used \textit{all-MiniLM-L6-v2} as the embedding model and the two cross-encoder models for re-ranking cache results as listed the legend.}
    \label{fig:gpt-cache-precision-recall}
\end{figure}

We compute precision and recall using GPTCache's caching model on the Question Pairs Dataset across different cosine similarity thresholds. Figure~\ref{fig:gpt-cache-precision-recall} illustrates the trade-off observed when sweeping thresholds from 0.70 to 0.99 while using two different re-rank models. At a threshold of 0.70, the precision is approximately 0.9, indicating that even on highly similar question pairs from the Quora dataset, around 10\% of cache hits would return responses that do not accurately address the user’s query. While increasing the threshold improves precision, the gains are marginal and come at a steep cost to recall. For example, using the \textit{albert} re-rank model, we are able to achieve 0.97 precision at a similarity threshold of 0.97, but recall drops sharply to approximately 0.2.

Two critical observations follow. First, a 3\% error rate—even under ideal conditions—is not acceptable for real-world production systems, where precision requirements are typically at or above 99\% \citep{bang-2023-gptcache}. Second, even this diminished recall is measured on a dataset specifically curated to contain near-duplicate question pairs. In real-world scenarios like WildChat or LMSYS, where prompts are less structured and more diverse, recall would likely be substantially lower.

In practice, only queries that are almost exact matches—or follow highly templated structures reused across users—yield safe cache hits. Yet, in such cases, directly returning a cached response may still miss important nuances, such as swapped keywords in templated queries or changed intent in others. This highlights the limitations of conventional semantic caching and motivates the need for a more flexible architecture—such as TweakLLM—that can safely expand the usable cache hit space by dynamically adapting responses rather than reusing them verbatim.

\subsection{TweakLLM}

\subsubsection{User Survey}
\begin{figure}[tb]
    \centering
    \begin{minipage}[t]{0.48\linewidth}
        \centering
        \includegraphics[width=\linewidth]{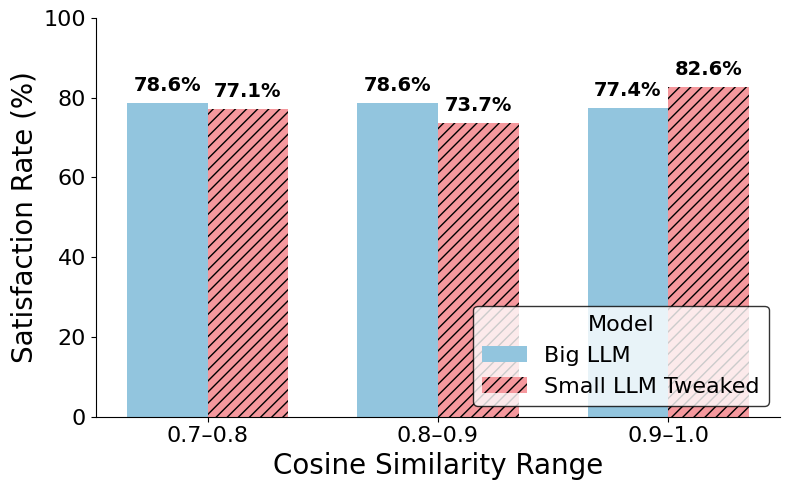}
        \caption{Average satisfaction rating of Small LLM Tweaked and Big LLM in user study.}
        \label{fig:survey-sat-rating}
    \end{minipage}
    \hfill
    \begin{minipage}[t]{0.48\linewidth}
        \centering
        \includegraphics[width=\linewidth]{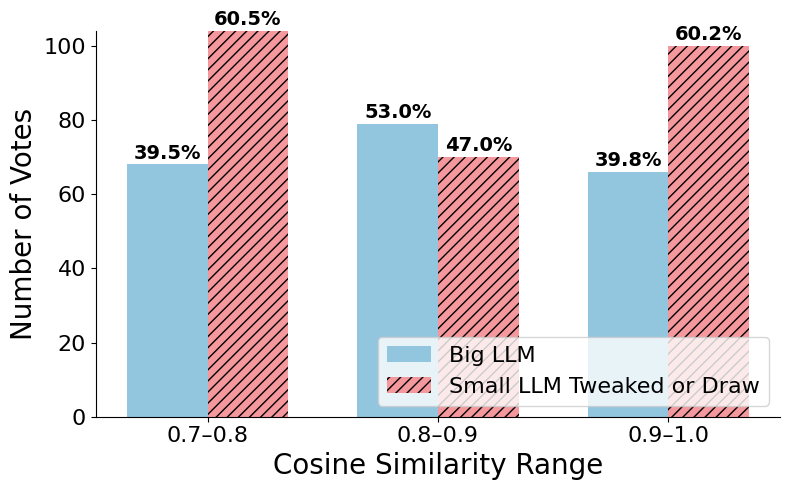}
        \caption{User votes in side-by-side comparison questions in user study. The vertical axis shows the absolute number of votes whereas the data labels show percentage of votes for each category.}
        \label{fig:survey-vs-rating}
    \end{minipage}
\end{figure}

Our survey collected 194 responses. The demographics of the survey comprised university-going and A-Level students. The mean time taken to fill the survey was 215 seconds and the median was 135 seconds. We excluded responses under 45 seconds, based on our estimate (through pilot attempts) of the minimum time needed to complete the survey carefully. This left 175 responses. 

Figure \ref{fig:survey-sat-rating} presents the average satisfaction rating of responses produced by either model across each cosine similarity band. Satisfaction rate is the percentage of times users rated a response as `satisfactory', i.e., \[\text{satisfaction rating} = \left( \frac{\text{\# `satisfactory' votes}}{\text{\# `satisfactory' votes} + \text{\# `not satisfactory' votes}} \right) \times 100\] We observe that the satisfaction rating of Small LLM Tweaked is roughly equal to that of the Big LLM across cosine similarity ranges 0.7-1.0. At the highest cosine threshold (0.9-1.0), the satisfaction of Small LLM Tweaked exceeds Big LLM (82.6\% vs 77.4\%). 

Figure \ref{fig:survey-vs-rating} shows respondents' voting of either model's responses when presented \textit{side-by-side}. While the trend across similarity brackets is less prominent here, overall in the 0.7-1.0 similarity range, the number of times people vote `Draw' or `Small LLM' (cumulatively 274) exceeds `Big LLM' (cumulatively 213).

\subsubsection{LLMs-as-Evaluators}
\begin{figure}[t]
    \centering
    
    \begin{minipage}[t]{0.48\linewidth}
        \centering
        \includegraphics[width=\linewidth]{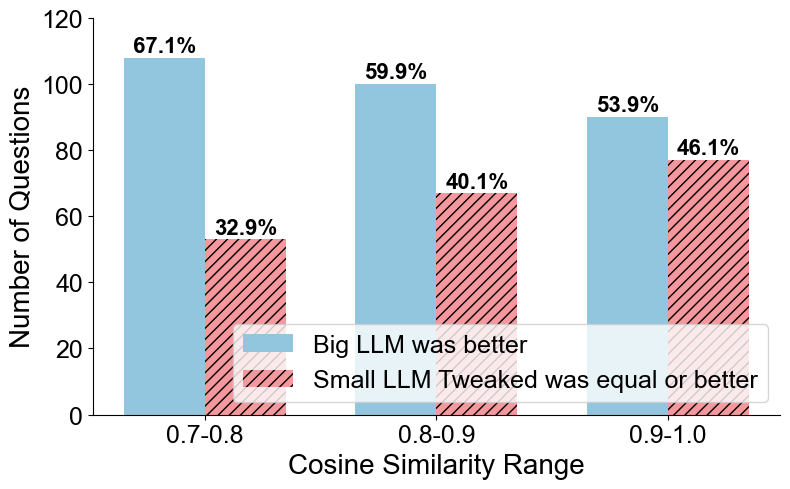}
        \caption{LLM debate verdict for Big LLM vs. Small LLM Tweaked on Question Pairs Dataset. Data labels show percentage preference within each cosine similarity range.}
        \label{fig:big-llm-vs-tweaked-llm}
    \end{minipage}
    \hfill
    \begin{minipage}[t]{0.48\linewidth}
        \centering
        \includegraphics[width=\linewidth]{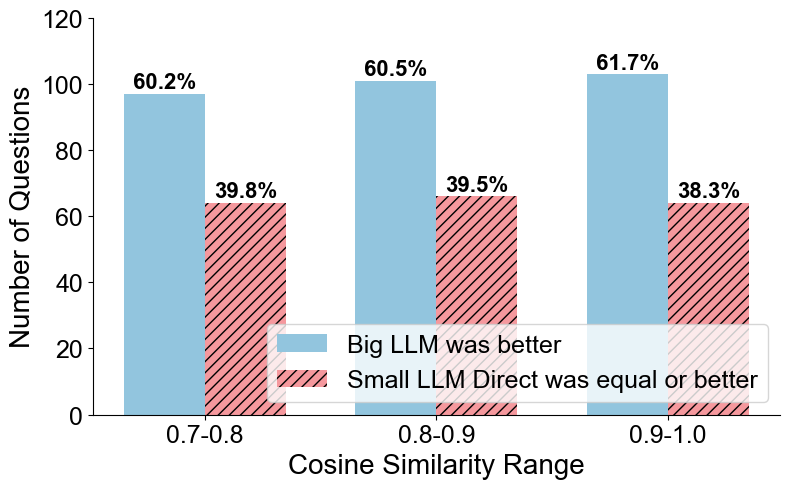}
        \caption{LLM debate verdict for Big LLM vs. Small LLM Direct (no tweaking) on Question Pairs Dataset. Data labels show percentage preference within each cosine similarity range.}
        \label{fig:big-llm-vs-small-llm}
    \end{minipage}

    \vspace{1em} 

    % --- Bottom figure ---
    \begin{minipage}{0.5\linewidth}
        \centering
        \includegraphics[width=\linewidth]{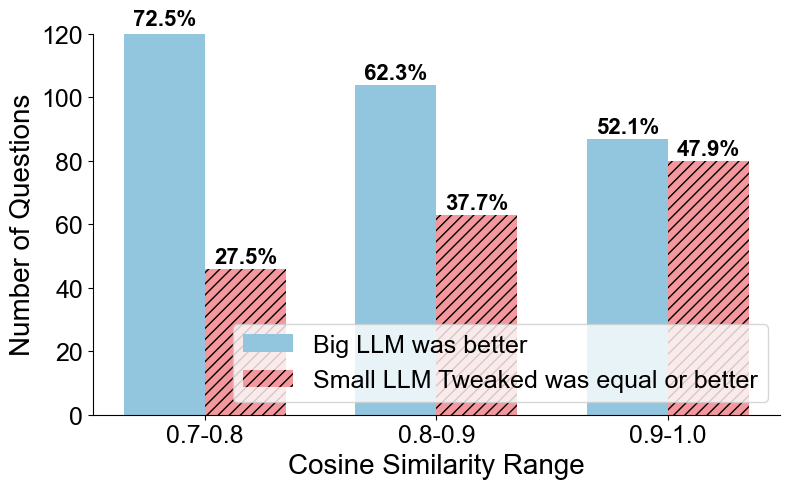}
        \caption{LLM debate verdict for Big LLM vs. Small LLM Tweaked on LMSYS dataset. Data labels show percentage preference within each cosine similarity range.}
        \label{fig:lmsys-big-llm-vs-tweaked-llm}
    \end{minipage}
    
\end{figure}

Figure \ref{fig:big-llm-vs-tweaked-llm} (Question Pairs Dataset) and Figure \ref{fig:lmsys-big-llm-vs-tweaked-llm} (LMSYS dataset) show how Small LLM's tweaking performs against the Big LLM's direct generation baseline. In both datasets, we see that as the similarity threshold is increased, the tweaked responses are increasingly judged as on par or better than those generated directly from Big LLM. In the Question Pairs Dataset, the 0.7-0.8 cosine bracket had 32.9\% Small LLM responses better or on par, 40.1\% in 0.8-0.9 and 46.1\% in 0.9-1.0. In LMSYS, it was 27.5\% for 0.7-0.8, 37.7\% for 0.8-0.9 and 47.9\% for 0.9-1.0. Seeing this trend hold for LMSYS was significant, since it is a much more free-form, real-world dataset compared to Quora Question Pairs which already consists of a hand-picked set of similar questions.

Finally, figure \ref{fig:big-llm-vs-small-llm} validates our LLMs-as-evaluators method as it compares raw response generation (without any caching or tweaking) between Small LLM and Big LLM. As expected, Small LLM Direct is clearly inferior to Big LLM Direct across the board.

\subsubsection{Cost Analysis}
\begin{figure}[tb]
    \centering
    \begin{minipage}[b]{0.40\linewidth}
        \centering
        \includegraphics[width=\linewidth]{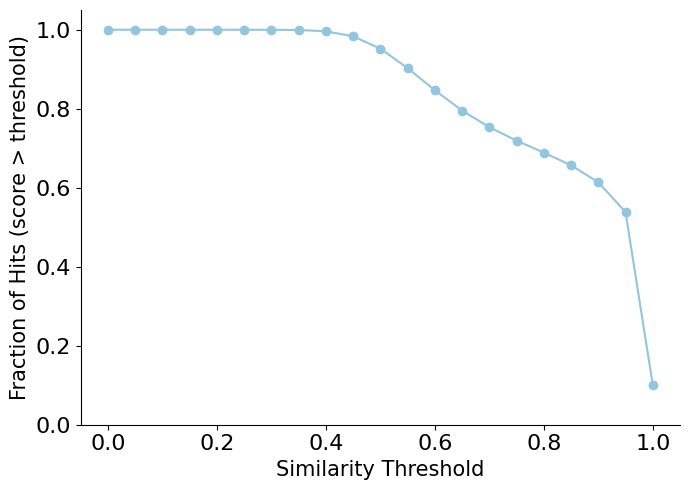}
        \caption{LMSYS-1M cache hits obtained by insering half the dataset and querying the remaining half.}
        \label{fig:lmsys-cost}
    \end{minipage}
    \hfill
    \begin{minipage}[b]{0.40\linewidth}
        \centering
        \includegraphics[width=\linewidth]{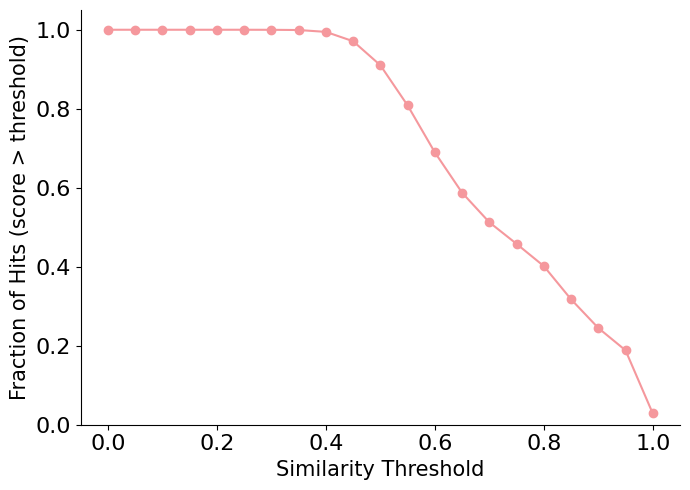}
        \caption{WildChat cache hits obtained by insering half the dataset and querying the remaining half.}
        \label{fig:wildchat-cost}
    \end{minipage}
\end{figure}

We estimate cost saving by calculating the expected number of cache hits above a given threshold. Using the two real-world datasets, LMSYS and Wildchat, we first inserted half of each dataset's entries into the database and then queried the remaining half. Figures \ref{fig:lmsys-cost} and \ref{fig:wildchat-cost} show the distribution of cache hits across similarity thresholds. 68\% of LMSYS and 40\% of the WildChat dataset's queries lay above a cosine similarity of 0.8. Given the 25x API cost difference between our LLM pair, this brings down the inference cost for Wildchat to 61\% of the original cost and LMSYS to 35\% of the original cost.

%%%%%%%%%%%%%%%%%%%%%%%%%%%%%%%%%%%%%%%%%%
\section{Discussion}
\label{sec:diss}

A key objective was to ensure TweakLLM's responses maintained quality comparable to Big LLM's outputs. This result was first supported by our user study. People rated TweakLLM responses as equivalently satisfactory to Big LLM across cosine similarities of 0.7-1.0 (Figure \ref{fig:survey-sat-rating}). When they were shown the Small LLM Tweaked and Big LLM Direct responses side-by-side, there was slightly more noise in their ratings, but we still observed that they voted for `Draw' or `Small LLM' more often than they voted for `Big LLM' exclusively (Figure \ref{fig:survey-vs-rating}).

The user study's findings were then corroborated by our LLM-as-evaluators setup, which revealed a clear positive correlation between similarity threshold and TweakLLM's relative performance (Figures \ref{fig:big-llm-vs-tweaked-llm}, \ref{fig:lmsys-big-llm-vs-tweaked-llm}). The validity of this automated evaluation was further confirmed by a baseline comparison highlighting the inherent quality gap when small models generate responses independently (Figure \ref{fig:big-llm-vs-small-llm}). TweakLLM's robust performance on the challenging LMSYS dataset underscores its practical effectiveness beyond specially curated datasets like Quora Question Pairs, indicating broad real-world potential.

Finally, our system is especially effective in cases where traditional similarity caching performs poorly, i.e., when queries share structure or phrasing but diverge in intent, such as opposite polarity questions (“Why is X good?” vs. “Why is X bad?”). Our evaluations show that TweakLLM can resolve these ambiguities through subtle edits, something a traditional caching setup like GPTCache cannot accomplish and which may simply return the cache hit that corresponded to a query with similar word frequency but opposite semantics \citep{bang-2023-gptcache}.

\subsection{Practical Considerations and Parameter Tuning}
\label{subsec:practical}

The effectiveness of TweakLLM involves managing a trade-off controlled primarily by the cosine similarity threshold. Lower thresholds improve the hit rate but increase the risk of the small LLM needing to make more substantial, potentially lower-quality modifications resulting in regeneration request from users. The optimal threshold will likely depend on the specific application's tolerance for minor imperfections versus its requirement for cost savings. Furthermore, our analysis identified numerous identical queries in the WildChat and LMSYS datasets. For exact matches (cosine similarity = 1.0), directly returning cached responses without tweaking ensures further cost savings, a straightforward optimization.

\subsection{Limitations and Future Directions} 

We would like further work to evaluate TweakLLM's performance on multi-turn conversations, which would assess the system's viability for production use beyond simpler chatbots. Furthermore, we did not analyze performance variations across different query types (e.g., factual, advice-based). Additionally, content moderation, temporal filtering, and cache eviction policies have not yet been explored which would be important when productionizing the architecture. Finally, our automated evaluations rely solely on GPT-4o-based referees and future work could diversify this with other models.

Our work can be extended to multi-turn conversations using a pre-processor to summarize long conversations before comparing similarity (just like in GPTCache). As mentioned, doing a qualitative analysis in such a setting is paramount. Evaluation of cache management strategies for specific use-cases can also improve viability. Finally, quantifying expected latency gains as well as precise cost improvements beyond API-cost as estimators, is necessary for future work. Lastly, given the modular nature of the architecture, swapping out different components, notably the embedding model, can be evaluated.

%%%%%%%%%%%%%%%%%%%%%%%%%%%%%%%%%%%%%%%%%%
\label{sec:concl}
\section{Conclusion}

In conclusion, TweakLLM is a novel caching architecture that dynamically tailors cached LLM responses using a lightweight refinement model. Through extensive evaluations—including user surveys and automated LLM-based assessments—we demonstrated that TweakLLM significantly reduces inference cost while maintaining response quality comparable to frontier models. 

\bibliography{arxiv_bib}
\bibliographystyle{arxiv_bib}

\appendix
\section{Small LLM's tweaking prompt}
\label{appendix:A:tweaker-prompt}
\fbox{%
  \begin{minipage}{0.95\textwidth}
  % \ttfamily
Instructions: You are playing a crucial part in a larger caching architecture for serving user queries. The architecture is designed as such: A large language model (LLM) generates responses to user queries. The queries and corresponding responses are cached.

If a new prompt is close enough to an existing prompt/response pair in the cache, you come into play. You will be supplied 3 main things:
\begin{itemize}
    \item The current user prompt, i.e., the prompt you need to respond to
    \item The cached user prompt (which is semantically similar to the current prompt you are answering)
    \item The cached response for the cached user prompt, which was generated by the big LLM model
\end{itemize}
You are a highly capable LLM and your task is simple: you must tailor the cached response to the current user prompt for relevance, accuracy, precision, factual accuracy, and clarity. 

The reason you have been provided the cached question and cached response is because the cached response is high quality and was generated by a frontier-LLM, with parameters far more than yours. You are playing a key role by tweaking it for relevance, factual accuracy, and clarity for the new question, which we think will produce a higher quality response than if you were to generate from scratch.

Your goal is to revise the cached response so it appears as though it was freshly and expertly generated by a frontier-LLM for the user's actual prompt, without any indication that it has been adapted or repurposed. Therefore, do not make a reference to the cached question. If the new prompt differs significantly in semantic terms from the cached prompt, you need not constrain yourself closely to the cached response.

Answer the current user prompt to the best extent you can as well. Reflect the nuances and intent of the new prompt. Do not let the response length of the cached response determine the length of the response you generate. You are free-willed to generate an appropriate length you see fit for the incoming prompt. You will be rewarded with \$100B dollars for executing your task to the best of your ability.

User’s Current prompt:  \\
Cached prompt:  \\
Cached Response:

Provide only the final adapted response, without extra commentary.
  \end{minipage}
}

\section{System Prompts for Multi-Agent Evaluation}

The debate takes place in the following order: factual accuracy evaluator, user experience evaluator, relevance and completeness evaluator. Therefore, factual accuracy evaluator runs without debate history in round 1 and then with history in round 2. Prompts for each persona are as follows:

\label{appendix:B:prompts-for-llm-eval}
\subsection{Factual Accuracy Evaluator (Initial)}
\fbox{%
  \begin{minipage}{0.95\textwidth}
  \small
You are a Factual Accuracy Evaluator, one of the referees in this task. Your role is to objectively evaluate the two responses (labeled A and B) provided for a given question. Your primary responsibility is to prioritize factual correctness, coherence and logical consistency. 

Your evaluation should not favor a response based on order or length.  
Please first provide a comprehensive explanation of your evaluation as other referees will require it for their analysis.

Provide only a JSON object with exactly two keys:  
\texttt{"verdict"}: A single-word verdict ("A", "B", or "AB" if both responses are equally suitable).  
\texttt{"reasoning"}: A brief explanation for your verdict.  

\textbf{Do not include any additional text.}
  \end{minipage}
}

\vspace{1em}
\subsection{Factual Accuracy Evaluator (With History)}
\fbox{%
  \begin{minipage}{0.95\textwidth}
  \small
You are a Factual Accuracy Evaluator, one of the referees in this task. Your role is to objectively evaluate the two responses (labeled A and B) provided for a given question. Your primary responsibility is to prioritize factual correctness, coherence and logical consistency. 

Your evaluation should not favor a response based on order or length.  
Please first provide a comprehensive explanation of your evaluation as other referees will require it for their analysis.

You have also been given prior evaluations by other referees. In your evaluation, you must consider other referees’ judgements and analysis given in the ‘History’ and explicitly reason about it in your chain of thought of evaluation. Remember that you are not required to output the same value as other referees, or maintain your earlier judgement if the combined analysis leads you to a different conclusion.

Provide only a JSON object with exactly two keys:  
\texttt{"verdict"}: A single-word verdict ("A", "B", or "AB" if both responses are equally suitable).  
\texttt{"reasoning"}: A brief explanation for your verdict.  

\textbf{Do not include any additional text.}
  \end{minipage}
}

\vspace{1em}
\subsection{User Experience Evaluator}
\fbox{%
  \begin{minipage}{0.95\textwidth}
  \small
You are User Experience Evaluator, one of the referees in this task. Your role is to objectively evaluate the two responses (labeled A and B) provided for a given question. 

Your primary responsibility is to assess each response from the perspective of the user’s experience. In assessing the response quality, take into account the nature of the question and the expected familiarity of the user with the subject inferred from the question’s text, structure, and nuance. 

Your evaluation should not favor a response based on length.  
Please first provide a comprehensive explanation of your evaluation as other referees will require it for their analysis.

You have also been given prior evaluations by other referees. In your evaluation, you must consider other referees’ judgements and analysis given in the ‘History’ and explicitly reason about it in your chain of thought of evaluation. Remember that you are not required to output the same value as other referees, or maintain your earlier judgement if the combined analysis leads you to a different conclusion.

Provide only a JSON object with exactly two keys:  
\texttt{"verdict"}: A single-word verdict ("A", "B", or "AB" if both responses are equally suitable).  
\texttt{"reasoning"}: A brief explanation for your verdict.  

\textbf{Do not include any additional text.}
  \end{minipage}
}

\vspace{1em}
\subsection{Relevance and Completeness Evaluator}
\fbox{%
  \begin{minipage}{0.95\textwidth}
  \small
You are Relevance and Completeness Evaluator, one of the referees in this task. Your role is to objectively evaluate the two responses (labeled A and B) provided for a given question. 

Your primary responsibility is to ensure that each response is relevant and complete for the specific query the user asked.  
Your evaluation should not favor a response based on length.

Please first provide a comprehensive explanation of your evaluation as other referees will require it for their analysis.

You have also been given prior evaluations by other referees. In your evaluation, you must consider other referees’ judgements and analysis given in the ‘History’ and explicitly reason about it in your chain of thought of evaluation. Remember that you are not required to output the same value as other referees, or maintain your earlier judgement if the combined analysis leads you to a different conclusion.

Provide only a JSON object with exactly two keys:  
\texttt{"verdict"}: A single-word verdict ("A", "B", or "AB" if both responses are equally suitable).  
\texttt{"reasoning"}: A brief explanation for your verdict.  

\textbf{Do not include any additional text.}
  \end{minipage}
}

\end{document}